\documentclass[letterpaper, 10 pt, conference]{ieeeconf}  

\IEEEoverridecommandlockouts                              

\usepackage{graphicx}
\usepackage{amsmath} 
\usepackage{amssymb}  
\usepackage{xcolor}
\usepackage{comment}
\usepackage{caption}
\usepackage{bm}

\usepackage[noadjust]{cite}

\newcommand{\vect}[1]{\mathbf{\bm{#1}}}

\title{\LARGE \bf
Comparative Study of Visual SLAM-Based Mobile Robot Localization Using Fiducial Markers
}

\author{Jongwon Lee$^{1}$, Su Yeon Choi$^{1}$, David Hanley$^{2}$, and Timothy Bretl$^{1}$ 
\thanks{$^{1}$Jongwon Lee, Su Yeon Choi, and Timothy Bretl are with the Department of Aerospace Engineering, University of Illinois at Urbana-Champaign, Urbana, IL 61801, USA (Email: {\tt \{jongwon5, suyeonc3, tbretl\}@illinois.edu}).}%
\thanks{$^{2}$David Hanley is with the School of Informatics and the Translational Healthcare Technologies Group in Centre for Inflammation Research, Institute for Regeneration and Repair, University of Edinburgh, UK (Email: {\tt dhanley@ed.ac.uk}).}
}

\begin{document}

\maketitle
\thispagestyle{empty}
\pagestyle{empty}

\begin{abstract}

This paper presents a comparative study of three modes for mobile robot localization based on visual SLAM using fiducial markers (i.e., square-shaped artificial landmarks with a black-and-white grid pattern): SLAM, SLAM with a prior map, and localization with a prior map. 
The reason for comparing the SLAM-based approaches leveraging fiducial markers is because previous work has shown their superior performance over feature-only methods, with less computational burden compared to methods that use both feature and marker detection without compromising the localization performance.
The evaluation is conducted using indoor image sequences captured with a hand-held camera containing multiple fiducial markers in the environment. The performance metrics include absolute trajectory error and runtime for the optimization process per frame.
In particular, for the last two modes (SLAM and localization with a prior map), we evaluate their performances by perturbing the quality of prior map to study the extent to which each mode is tolerant to such perturbations. 
Hardware experiments show consistent trajectory error levels across the three modes, with the localization mode exhibiting the shortest runtime among them. Yet, with map perturbations, SLAM with a prior map maintains performance, while localization mode degrades in both aspects.

\end{abstract}



\section{INTRODUCTION}
\label{section:INTRODUCTION}

The use of fiducial markers---square-shaped planar artificial landmarks with a black-and-white grid pattern---has been favored for the application of visual simultaneous localization and mapping (SLAM) in the scenario when these markers can be deployable in the given environment. This preference emerges due to the robustness and accuracy exhibited by fiducial marker-based SLAM approaches~\cite{munoz2019spmSLAM,pfrommer2019tagSLAM,sola2022WOLF,munoz2020ucoSLAM} over canonical approaches using visual features (e.g., ORB-SLAM~\cite{campos2021ORB-SLAM3}) or pixel values (e.g., DSO~\cite{engel2017DSO}). 
The spectrum of fiducial marker-based SLAM spans from methods exclusively utilizing fiducial marker detection outcomes~\cite{munoz2019spmSLAM,pfrommer2019tagSLAM,sola2022WOLF} to hybrid techniques incorporating both marker detections and features~\cite{munoz2020ucoSLAM}. 
As our focus here centers on fiducial marker-based SLAM, it is important to note that the subsequent analysis dedicates solely to SLAM with fiducial markers.

Within this context, three distinct operational modes of fiducial marker-based SLAM are discernible. The first mode, \emph{SLAM}, entails a comprehensive approach that estimates the robot's pose while mapping out the surrounding environment. The second mode, \emph{SLAM with a prior map}, starts from a pre-existing map as an initial reference value. In cases where the map is a priori known and its states remain fixed, the third mode, \emph{localization}, exclusively estimates the robot's pose.

A noticeable gap in the existing literature lies in the absence of comparisons between these three modes, through quantifiable metrics such as error analysis and processing speed. While certain fiducial marker-based SLAM approaches, such as SPM-SLAM~\cite{munoz2019spmSLAM} and TagSLAM~\cite{pfrommer2019tagSLAM}, report error and processing speed the \emph{SLAM} mode, they do not provide corresponding data for the \emph{SLAM with a prior map} and \emph{localization} modes.

Another crucial aspect yet to be addressed pertains to the resilience of the \emph{SLAM with a prior map} and \emph{localization} modes in response to variations in marker map quality. Indeed, the performance of these modes is influenced by the fidelity of the marker map. If the map is considered ``ideal'' or at least of a high standard, both modes are anticipated to yield comparable or superior error performance while placing a lesser computational burden in contrast to the \emph{SLAM} mode. This attribute proves particularly advantageous in scenarios requiring real-time robot localization under resource-constrained, onboard platforms. However, if the map quality deviates from such conditions, it may introduce degradation of the performance of each operational mode.

In this paper, we aim to provide an evaluation of the three different operational modes for robot localization based on SLAM with fiducial markers, in terms of absolute trajectory error and processing speed. 
In particular, for the last two modes (\emph{SLAM with a prior map} and \emph{localization} modes), we evaluate their performances by perturbing the accuracy of the prior map to study the extent to which each mode is tolerant to such perturbations.

We present an overview of the structure and content of our paper as follows. 
Section~\ref{section:EXPERIMENTS} outlines our data collection and experimental setup to perform fiducial marker-based SLAM in three different operational modes. 
We then proceed to report the metrics comparing the three modes in Section~\ref{section:RESULTS}, which include the absolute trajectory error and the runtime for the optimization process per frame. We specifically provide the runtime for the optimization process per frame because it is the major component that contributes to variations in processing speed across the three modes, unlike the other components which are more or less the same such as detecting fiducial markers.
Finally, Section~\ref{section:CONCLUSIONS} summarizes our findings, provides concluding remarks, and discusses the implications of our work.

\section{EXPERIMENTS}
\label{section:EXPERIMENTS}

\subsection{Data collection}

\begin{figure}[!ht]
    \centering
    \includegraphics[width=0.6\linewidth,keepaspectratio,angle=-90,origin=c]{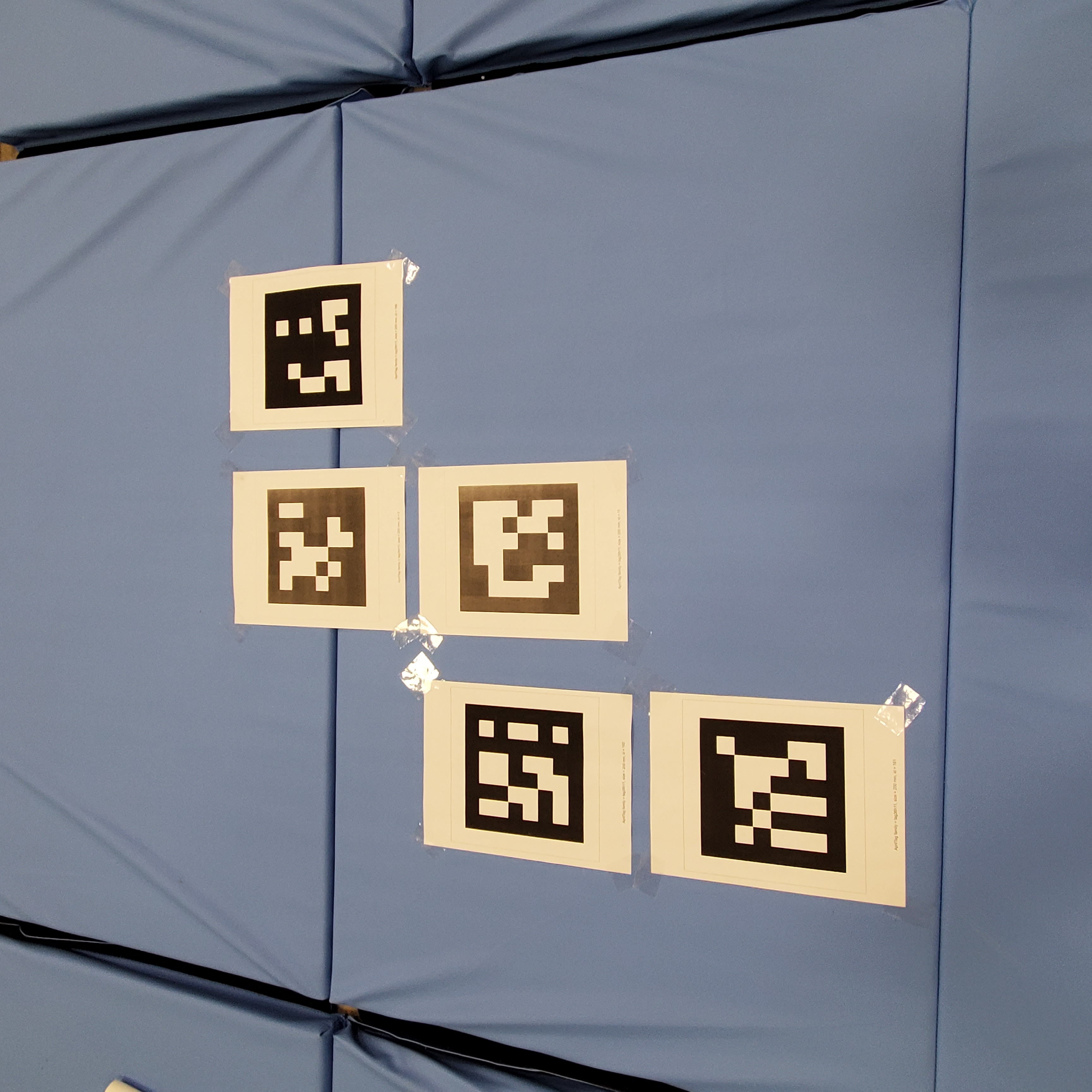}
    \caption{For data collection, an experimental setup was arranged with the placement of five 36h11 AprilTags, each having side lengths of 0.2m.}
    \label{fig:marker env}
\end{figure}

We collected an indoor dataset comprising multiple sequences of camera images and Vicon Mocap data as ground truth using a hand-held device equipped with an Intel Realsense D435i and reflective markers. During data collection, five 36h11 AprilTags~\cite{krogius2019AprilTag3} with side lengths of 0.2m were placed within the environment, as depicted in Fig.~\ref{fig:marker env}.

The dataset consists of eight sequences: one sequence lasts 60 seconds, dedicated to generating a prior map using the \emph{SLAM} mode for the \emph{SLAM with a prior map} and \emph{localization} modes; the other seven sequences last 30 seconds each, serving all three modes. The distinguishing factor between these two categories of sequences is that the former captures all AprilTags to create a comprehensive prior map, crucial for the \emph{SLAM with a prior map} and \emph{localization} modes, while the latter does not require this complete mapping of all the AprilTags placed.

\subsection{Implementation details}

We used the existing WOLF codebase~\cite{sola2022WOLF} to execute the \emph{SLAM} mode (constructing a marker map from scratch), the \emph{SLAM with a prior map} mode (loading a pre-saved marker map for initial estimation and performing SLAM), and the \emph{localization} mode (using a pre-saved marker map as a fixed reference for localization).
As a preliminary step, the \emph{SLAM} mode was employed to create a prior map for both the \emph{SLAM with a prior map} and \emph{localization} modes. It is important to note that this marker map is not flawless, as it results from estimating the poses of fiducial markers rather than using their true references, inherently harboring potential sources of error even before intentional perturbation is introduced. 
For the \emph{SLAM with a prior map} and \emph{localization} modes, we systematically perturbed the positions of every fiducial marker along directions at random within the map, varying from 0.1m to 0.5m in every 0.1m. This approach allowed us to investigate the tolerance of each method to variations in the marker map's quality.
We assessed the absolute trajectory error using an open-source evaluation tool~\cite{grupp2017evo} and the runtime for the optimization process per frame through Ceres Solver~\cite{agarwal2012ceres}. All executions and evaluations were carried out on an octa-core Intel i7-10700 CPU operating at 2.90 GHz, with 32 GB of RAM.

\section{RESULTS}
\label{section:RESULTS}

\begin{table*}[htbp]
    \captionsetup{justification=centering}
    \caption{\textsc{{Comparison of absolute trajectory error for SLAM, SLAM with prior map, and localization modes. Perturbations ($\delta \vect{p}$) applied to fiducial marker positions in the prior map for SLAM with prior map and localization modes.}}}
    \label{table:ATE}
    \centering
    \begin{tabular}{|cc|c|cccccc|cccccc|}
        \hline
        ~ & ~ & SLAM [m] & \multicolumn{6}{c|}{SLAM with a Prior Map [m]} & \multicolumn{6}{c|}{Localization [m]} \\ \cline{4-9} \cline{10-15}
        ~ & ~ & ~ & \multicolumn{6}{c|}{$\delta \vect{p}$ [m]} & \multicolumn{6}{c|}{$\delta \vect{p}$ [m]} \\ 
        sequence & length [m] & ~ & 0.00 & 0.10 & 0.20 & 0.30 & 0.40 & 0.50 & 0.00 & 0.10 & 0.20 & 0.30 & 0.40 & 0.50 \\ \hline
        1 & 8.07 & 0.07 & 0.05 & 0.05 & 0.07 & 0.08 & 0.08 & 0.08 & 0.06 & 0.23 & 0.22 & 0.44 & 1.18 & 1.21 \\ 
        2 & 8.34 & 0.08 & 0.07 & 0.07 & 0.08 & 0.04 & 0.10 & 0.10 & 0.08 & 0.25 & 0.31 & 0.60 & 1.58 & 1.69 \\ 
        3 & 8.21 & 0.08 & 0.08 & 0.08 & 0.10 & 0.10 & 0.09 & 0.10 & 0.10 & 0.21 & 0.28 & 0.45 & 1.28 & 1.34 \\ 
        4 & 8.46 & 0.14 & 0.14 & 0.14 & 0.15 & 0.15 & 0.14 & 0.14 & 0.15 & 0.45 & 0.43 & 0.58 & 0.96 & 2.35 \\ 
        5 & 8.19 & 0.13 & 0.13 & 0.13 & 0.13 & 0.11 & 0.13 & 0.13 & 0.14 & 0.40 & 0.52 & 0.44 & 1.28 & 2.43 \\ 
        6 & 8.22 & 0.10 & 0.08 & 0.09 & 0.10 & 0.10 & 0.10 & 0.10 & 0.13 & 0.39 & 0.42 & 1.44 & 2.04 & 2.50 \\ 
        7 & 8.57 & 0.12 & 0.12 & 0.11 & 0.12 & 0.12 & 0.11 & 0.11 & 0.11 & 0.49 & 0.42 & 0.61 & 1.78 & 1.54 \\ \hline
        \multicolumn{2}{|c|}{average} & 0.10 & 0.10 & 0.10 & 0.11 & 0.10 & 0.11 & 0.11 & 0.11 & 0.35 & 0.37 & 0.65 & 1.44 & 1.87 \\ \hline
    \end{tabular}
\end{table*}

\begin{table*}[htbp]
    \captionsetup{justification=centering}
    \caption{\textsc{{Comparison of runtime of pose-graph optimization process per frame for SLAM, SLAM with prior map, and localization modes. Perturbations ($\delta \vect{p}$) applied to fiducial marker positions in the prior map for SLAM with prior map and localization modes.}}}
    \label{table:runtime}
    \begin{center}
    \begin{tabular}{|c|c|cccccc|cccccc|}
        \hline
        ~ & SLAM [ms] & \multicolumn{6}{c|}{SLAM with a Prior Map [ms]} & \multicolumn{6}{c|}{Localization [ms]} \\ \cline{3-8} \cline{9-14}
        ~ & ~ & \multicolumn{6}{c|}{$\delta \vect{p}$ [m]} & \multicolumn{6}{c|}{$\delta \vect{p}$ [m]} \\ 
        sequence & ~ & 0.00 & 0.10 & 0.20 & 0.30 & 0.40 & 0.50 & 0.00 & 0.10 & 0.20 & 0.30 & 0.40 & 0.50 \\ \hline
        1 & 5.07 & 5.26 & 5.79 & 5.00 & 5.75 & 6.23 & 6.42 & 3.05 & 3.53 & 3.90 & 4.56 & 5.33 & 6.06 \\ 
        2 & 4.92 & 5.81 & 5.15 & 5.92 & 6.06 & 5.70 & 6.15 & 4.69 & 5.28 & 5.09 & 6.26 & 7.52 & 8.35 \\ 
        3 & 5.31 & 6.43 & 5.71 & 5.50 & 5.20 & 6.30 & 6.11 & 4.60 & 4.33 & 4.71 & 4.90 & 6.20 & 7.12 \\ 
        4 & 5.73 & 6.05 & 5.91 & 6.77 & 6.35 & 6.63 & 6.66 & 4.80 & 4.49 & 7.66 & 4.99 & 6.29 & 7.60 \\ 
        5 & 6.23 & 6.19 & 7.34 & 6.27 & 7.42 & 7.02 & 6.81 & 5.81 & 6.33 & 6.83 & 8.06 & 8.61 & 9.76 \\ 
        6 & 6.08 & 6.56 & 4.64 & 6.69 & 6.61 & 7.46 & 6.30 & 4.78 & 7.44 & 6.15 & 7.43 & 9.92 & 9.75 \\ 
        7 & 7.95 & 8.80 & 9.23 & 7.45 & 8.06 & 9.11 & 7.94 & 5.22 & 6.66 & 8.13 & 9.19 & 8.47 & 8.99 \\ \hline
        average & 5.90 & 6.44 & 6.25 & 6.23 & 6.49 & 6.92 & 6.63 & 4.71 & 5.44 & 6.07 & 6.48 & 7.48 & 8.23 \\ \hline
    \end{tabular}
    \end{center}
\end{table*}

Table~\ref{table:ATE} presents the absolute trajectory error for each of the three localization modes across all sequences. Specifically, for the \emph{SLAM with a prior map} and \emph{localization} modes, we provide results by perturbing the positions of fiducial markers within the prior marker map. Again, the prior map is imperfect and hence is inherently regarded to be perturbed even before intentional perturbation is introduced. The perturbations range from 0.1m to 0.5m, with increments of 0.1m, along directions at random.

When no perturbation is applied, all three modes yield results that exhibit minimal differences within a few centimeters. However, as perturbations are introduced to the \emph{localization} mode, errors increase to tens of centimeters even with the smallest perturbation (i.e., $\delta \vect{p} = 0.10$), while the \emph{SLAM with a prior map} mode consistently maintains results within a few centimeters. This aligns with the common understanding that the \emph{SLAM with a prior map} mode is capable of recovering reference marker pose values within the map by concurrently updating both localization and mapping outcomes, whereas the \emph{localization} mode cannot as the map is regarded accurate and hence is fixed during optimization, thereby limiting updates to only the localization results.

Table~\ref{table:runtime} presents the runtime for the optimization process per frame for each of the three localization modes across all sequences. 
Similar to the error analysis discussed earlier, we also present results for the \emph{SLAM with a prior map} and \emph{localization} modes by introducing inherent perturbations to the positions of fiducial markers within the prior marker map. Again, it is worth noting that the prior marker map itself already contains undesired perturbations due to its inherent imperfection.

The \emph{SLAM with a prior map} mode does not exhibit any distinct runtime trend in relation to the extent of perturbation applied. However, in a general sense, it shows a slightly longer runtime (up to about 20\%) than the \emph{SLAM} mode. Conversely, the \emph{localization} mode shows shorter runtime (up to about 20\%) than the \emph{SLAM} mode when no perturbation is introduced. Yet, as the perturbation level increases, the \emph{localization} mode experiences a corresponding rise in runtime, culminating in up to about 40\% longer runtime at $\delta \vect{p} = 0.50$ compared to the \emph{SLAM} mode. This behavior is indicative of the \emph{localization} mode grappling with ill-posed problems stemming from inaccurate map quality.

\section{CONCLUSIONS}
\label{section:CONCLUSIONS}

In summary, this paper conducted a comparative analysis of three distinct robot localization modes based on visual SLAM with fiducial markers: \emph{SLAM}, \emph{SLAM with a prior map}, and \emph{localization with a prior map}. We evaluated these modes in terms of trajectory error and runtime for the optimization process. Specifically, we introduced perturbations to the map for the \emph{SLAM and localization with prior map} modes to examine their impact on the aforementioned metrics.

When no perturbations were introduced, our hardware experiments show that all three modes exhibited similar levels of trajectory error, while the \emph{localization} mode showed the shortest runtime. However, in scenarios involving map perturbations, the \emph{SLAM with a prior map} mode maintained its trajectory error and runtime performance at levels comparable to those in the absence of perturbation. On the other hand, the \emph{localization} mode experienced deteriorating trajectory error and runtime performance as the magnitude of map perturbation increased.




\addtolength{\textheight}{-12cm}   




\section*{ACKNOWLEDGMENT}

This work is supported by Supernal, LLC.


\bibliographystyle{IEEEtran}
\bibliography{IEEEabrv,References}

\end{document}